\documentclass{article}
\usepackage{tencent_tech_report}
\usepackage[colorlinks = true,
            linkcolor = blue,
            urlcolor  = blue,
            citecolor = blue,
            anchorcolor = blue]{hyperref}

\usepackage{tcolorbox} 
\definecolor{tsinghuapurple}{RGB}{102,8,116}
\newtcolorbox{alprompt}[1]{
        boxrule = 1pt,
        fontupper = \small\tt,
        fonttitle = \bf\color{black},
        arc = 2pt,
        rounded corners,
        colframe = black,
        colbacktitle = white!97!yellow,
        colback = white!97!yellow,
        title = #1,
}

\usepackage{amssymb}
\usepackage{microtype}
\usepackage{hyperref}
\usepackage{url}
\usepackage{booktabs}
\usepackage{enumitem}
\usepackage{multicol}
\usepackage{multirow}
\usepackage{CJKutf8}
\usepackage{amsmath}
\usepackage{siunitx}
\usepackage{floatflt}
\usepackage{wrapfig}
\usepackage{graphicx}
\usepackage{booktabs}
\usepackage{wrapfig}
\usepackage{authblk}
\usepackage{lipsum}

\usepackage{algorithm}
\usepackage{algorithmicx}
\usepackage{algpseudocode}
\usepackage{microtype}
\usepackage{graphicx}
\usepackage{multirow}
\usepackage{booktabs} 
\usepackage{pifont}  
\usepackage{graphicx}  
\usepackage{subcaption} 
\usepackage{hyperref}

\usepackage{amssymb} 

\algnewcommand{\LeftComment}[1]{\Statex \(\triangleright\) #1}

\newtcolorbox{promptbox}[3][Prompt]{
colback=black!5!white,
arc=5pt, 
boxrule=0.5pt,
fonttitle=\bfseries,
title=#1, 
before upper={\small}, fontupper=\fontfamily{ptm}\selectfont,
colframe=#2,
label=#3,
}
\usepackage{array}
\usepackage{amsmath}
\usepackage{amssymb}
\usepackage{mathtools}
\usepackage{amsthm}
\usepackage{arydshln}
\usepackage[capitalize,noabbrev]{cleveref}
\usepackage{adjustbox} 
\usepackage{enumitem}
\usepackage{xspace}
\theoremstyle{plain}

\theoremstyle{definition}

\theoremstyle{remark}

\usepackage{xcolor}
\usepackage{tcolorbox}
\tcbuselibrary{listings,skins}

\definecolor{promptbg}{RGB}{245, 245, 245}   
\definecolor{promptborder}{RGB}{200, 200, 200} 
\tcbset{
    promptstyle/.style={
        enhanced,
        frame hidden,
        boxrule=0pt,
        left=8pt,
        right=8pt,
        top=6pt,
        bottom=6pt,
        arc=3pt,
        colback=promptbg,
        colframe=promptborder,
        fontupper=\fontsize{0.7em}{0.7em}\selectfont\ttfamily\leftskip,
        before upper={\parindent}, 
        overlay unbroken and first={
            \draw[promptborder, line width=1pt] 
                (frame.south west) -- (frame.north west);
            \draw[promptborder, line width=1pt] 
                ([xshift=-2pt]frame.north east) -- (frame.south east);
        },
        overlay middle and last={
            \draw[promptborder, line width=1pt] 
                (frame.south west) -- (frame.north west);
            \draw[promptborder, line width=1pt]
                ([xshift=-2pt]frame.north east) -- (frame.south east);
        }
    }
}

\usepackage[textsize=tiny]{todonotes}

\newcommand{\ignore}[1]{}

\colmfinalcopy

\definecolor{nred}{RGB}{196, 38, 11}
\definecolor{ngreen}{RGB}{18, 141, 21}
\definecolor{nblue}{RGB}{41, 52, 190}
\definecolor{norange}{RGB}{230, 106, 53}

\newcommand{\method}[0]{\textsc{BatonVoice}}
\newcommand{\model}[0]{\textsc{BatonTTS}}

\colmfinalcopy

\title{\method{}: An {\bf \em \color{ngreen}  Operationalist} Framework for Enhancing Controllable Speech Synthesis with Linguistic Intelligence from LLMs}

\author[ ]{\bf Yue Wang$^{1,2}$}
\author[ ]{\bf Ruotian Ma$^{1}$}
\author[ ]{\bf Xingyu Chen$^{1}$}
\author[ ]{\bf Zhengliang Shi$^{1}$}
\author[ ]{\bf Wanshun Chen$^{1}$}
\author[ ]{\bf Huang Liu$^{1}$}
\author[ ]{\\\bf Jiadi Yao$^{1}$}
\author[ ]{\bf Qu Yang$^{1}$}
\author[ ]{\bf Qingxuan Jiang$^{1}$}
\author[ ]{\bf Fanghua Ye$^{1}$}
\author[ ]{\bf Juntao Li~$^{*, 2}$}
\author[ ]{\bf Min Zhang~$^{2}$}
\author[ ]{\\\bf Zhaopeng Tu\thanks{Correspondence to: Zhaopeng Tu \textless zptu@tencent.com\textgreater~ and Juntao Li \textless ljt@suda.edu.cn\textgreater.}~~$^{,1}$}
\author[ ]{\bf Xiaolong Li$^{1}$}
\author[ ]{\bf Linus$^{1}$}

\affil[1]{Tencent Multimodal Department}
\affil[2]{Soochow University \protect\\[2pt] 
\url{https://github.com/Tencent/digitalhuman/tree/main/BatonVoice}}

\begin{document}

\maketitle

\vspace{-15pt}
\begin{quote}
``{\em In general, we mean by any concept nothing more than a set of operations.}''
\begin{flushright}
--- P. W. Bridgman
\end{flushright}
\end{quote}

\begin{figure}[h!]
\centering
\includegraphics[width=0.9\linewidth]{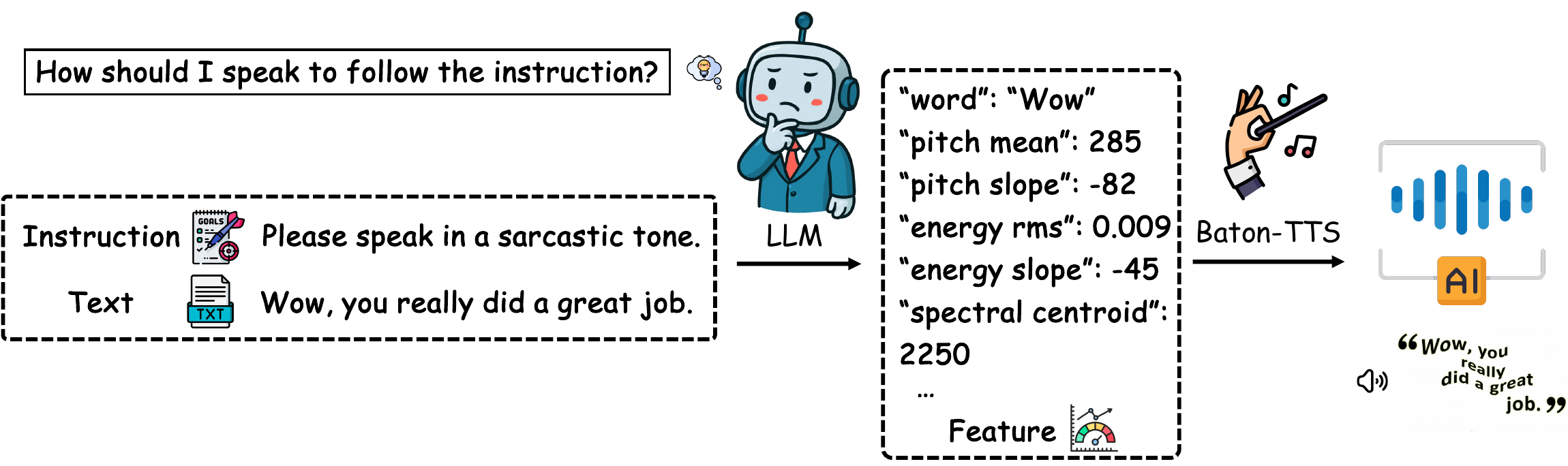}
\caption{Illustration of \textbf{\method{}}: (1) An LLM, acting as a \textbf{conductor}, interprets the user's instructions and generates explicit vocal features. (2) These features are then fed into \model{} model, the \textbf{orchestra}, which synthesizes the final speech. This separation allows the LLM to leverage its linguistic intelligence to guide the synthesis process, enabling controllable TTS.}
\label{fig:frontpage}
\end{figure}

\begin{abstract}
The rise of Large Language Models (LLMs) is reshaping multimodel models, with speech synthesis being a prominent application. However, existing approaches often underutilize the linguistic intelligence of these models, typically failing to leverage their powerful instruction-following capabilities. This limitation hinders the model's ability to follow text instructions for controllable Text-to-Speech~(TTS). To address this, we propose a new paradigm inspired by ``operationalism'' that decouples instruction understanding from speech generation. We introduce \method{}, a framework where an LLM acts as a ``conductor'', understanding user instructions and generating a textual ``plan'' -- explicit vocal features (e.g., pitch, energy). A separate TTS model, the ``orchestra'', then generates the speech from these features. To realize this component, we develop \model{}, a TTS model trained specifically for this task. Our experiments demonstrate that \method{} achieves strong performance in controllable and
emotional speech synthesis, outperforming strong open- and closed-source baselines. Notably, our approach enables remarkable zero-shot cross-lingual generalization, accurately applying feature control abilities to languages unseen during post-training. This demonstrates that objectifying speech into textual vocal features can more effectively unlock the linguistic intelligence of LLMs.
\end{abstract}

\section{Introduction}

The rapid advancement of Large Language Models (LLMs) has catalyzed a paradigm shift in Multimodal Large Language Models~(MLLMs), with frameworks now unifying text, images, and speech within a single model~\citep{zeng2024glm,xu2025qwen3,deng2025emerging}. In Text-to-Speech (TTS), this has led to a new generation of systems that fine-tune a pre-trained LLM as a backbone to generate speech~\citep{wang2023neuralcodeclanguagemodels,guo2023prompttts,zhang2025minimax}. However, a critical yet underexplored question remains: {\em Are we fully leveraging the linguistic intelligence of LLMs in these TTS models?}

Existing LLM-based TTS models primarily treat the LLM as a backbone. This approach typically involves designing a tokenizer to convert speech into discrete tokens and then training the model on large-scale datasets tailored to specific objectives. For instance, training a controllable TTS model necessitates extensive manual annotation of existing speech data to acquire the corresponding control labels and instructions~\citep{du2024cosyvoice,du2024cosyvoice2scalablestreaming}, a process that is not only prohibitively expensive but also suffers from low inter-annotator agreement. We contend that this methodology largely bypasses the LLM's inherent linguistic intelligence, such as its strong capabilities for complex context understanding and instruction following.

To address this, we draw inspiration from the principle of ``operationalism'', where complex concepts are understood through quantifiable, interpretable operations. For instance, to analyze imperceptible ultrasound, we use sensors to extract quantifiable features like frequency and amplitude. We posit that controllable TTS can be transformed by operationalizing user instructions into the desired vocal features. This reframes the problem: the LLM first leverages its linguistic intelligence to understand instructions and generate explicit vocal features, which then serves as input for a subsequent TTS model. This approach allows us to circumvent the need for manually annotating speech with controllable labels.

To realize this vision, we introduce \method{}, a novel TTS framework that decouples instruction understanding from speech generation, as illustrated in Figure~\ref{fig:frontpage}. \method{} employs an LLM as a ``conductor'', which interprets the user's instructions to explicit vocal features, like pitch and energy. This plan is then fed into a separate TTS model, the ``orchestra'', which generates the final speech. The ``orchestra'' in our framework is \model{}, a TTS model we trained specifically to synthesize high-quality speech conditioned on these textual vocal features.

Our experiments validate the power of this decoupled approach. \method{} achieves strong performance in emotional speech synthesis, outperforming strong open- and closed-source models. For example, our 1.7B parameter model achieves an emotion accuracy of 57.6\%, significantly surpassing all baselines. To verify our hypothesis, we show that stronger linguistic intelligence directly translates to superior synthesis: upgrading the ``conductor'' LLM from our 1.7B model to the more capable \textit{Gemini 2.5 Pro} boosts the final model's emotion accuracy from 29.8\% to 57.6\%. Furthermore, \method{} exhibits remarkable zero-shot cross-lingual generalization, accurately applying feature control abilities to Chinese, which is an unseen language during feature control training stage. This work not only advances controllable speech synthesis but also presents a promising new paradigm for MLLM development, demonstrating how objectifying modalities into text can more fully unlock the linguistic intelligence of LLMs.

Our contributions are three-fold:
\begin{itemize}[leftmargin=15pt]
\item We propose a novel paradigm for controllable speech synthesis, inspired by ``operationalism'', which decouples linguistic intelligence from speech generation via quantifiable, interpretable vocal features.
\item We present a methodology for realizing this paradigm, including a novel data pipeline that automatically generates instruction-feature pairs, and we introduce \model{}, a specialized TTS model trained on this data to generate speech from the vocal features.
\item Through extensive experiments, we demonstrate that our framework, \method{}, achieves strong performance in controllable, expressive speech synthesis. It exhibits superior emotional control and remarkable zero-shot cross-lingual generalization performance, validating the effectiveness of our operationalism-inspired approach.
\end{itemize}

\section{\method{}: A Framework for Controllable TTS}
In this section, we introduce \method{}, a controllable TTS framework capable of synthesizing speech that adheres to arbitrary text-based instructions. Adopting an operationalist stance, \method{} leverages LLMs to interpret users' instructions into a JSON list of fine-grained vocal features. The core of this framework is \model{}, a TTS model trained specifically developed to synthesize speech from these features. We first describe the overall framework and its inference process, followed by a detailed introduction of the \model{} architecture and its three-stage training pipeline.

\subsection{Overall Framework and Inference Process}
The inference process of the \method{} framework is structured in two stages. In the first stage, for a given input text $X$ and a corresponding instruction $I$, an external LLM~(specifically, \textit
{Gemini 2.5 Pro}) is employed to interpret the instruction. This interpretation yields a set of fine-grained vocal features, denoted as $F_v$. These features constitute a quantitative vocal plan and encompass the following attributes:

\begin{itemize}[leftmargin=12pt]
     \item \textbf{Pitch} ({\em mean} and {\em slope}): The average fundamental frequency and the overall intonational contour.
     \item \textbf{Energy} ({\em RMS} and {\em slope}): The average signal amplitude and its dynamic variations.
     \item \textbf{Timbre} ({\em spectral centroid}): The perceived brightness of the speech.
\end{itemize}
We provide the prompt template utilized for this feature prediction in the Appendix. Subsequently, in the second stage, this feature list $F_v$, along with the original text $X$, is fed into \model{} to synthesize the final speech.

\begin{figure}[t]
\centering
\includegraphics[width=0.8\linewidth]{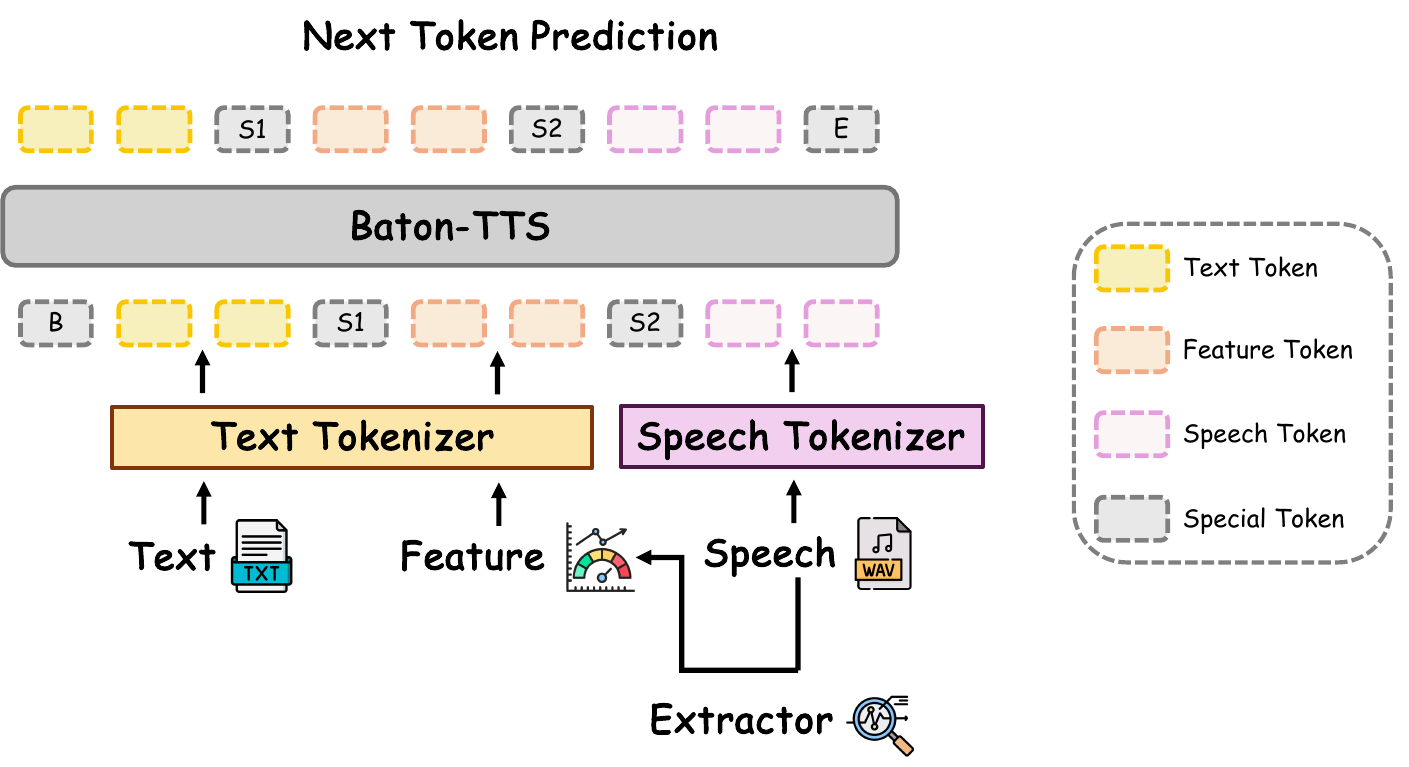}
\caption{Overview of the SFT stage of the \model{} framework. We extract vocal features from speech and verbalize them into a textual format.}
\label{fig:framework}
\end{figure}

\label{sec:framework}

\subsection{Model Architecture of \model{}}
We now detail the architecture of \model{}, the model responsible for generating speech from the specified feature list. Inspired by recent advancements such as CosyVoice2 \citep{du2024cosyvoice}, the architecture of \model{} comprises two primary components: an LLM backbone and a pre-trained speech decoder.

For the LLM backbone, we employ representative open-source models, specifically Qwen3-1.7B and Qwen2.5-0.5B. As will be demonstrated in our experimental section, our proposed method is effective across LLM backbones of varying capacities. The LLM is tasked with autoregressively generating a sequence that includes the input text to be synthesized, the corresponding speech features (i.e., the vocal plan), and the discrete speech tokens that realize this plan. The structure of this input sequence during the Supervised Fine-Tuning (SFT) stage is illustrated in Figure \ref{fig:framework}. It is important to note that while the features are part of the training sequence, during inference, they are generated by an external LLM as previously described.

For the final synthesis step, we leverage the speech decoder from the publicly available CosyVoice2 model. This decoder converts the discrete speech tokens produced by the LLM into a high-quality speech. It consists of a speech token encoder, a conditional flow matching model, and a HiFi-GAN vocoder. The flow matching model generates Mel spectrograms conditioned on the discrete speech tokens, and the HiFi-GAN vocoder then converts these spectrograms into the final speech. By utilizing a pre-trained speech decoder, we can focus our training efforts exclusively on teaching the LLM to control speech features through language. Consequently, the speech decoder remains frozen throughout our training process.

\subsection{Three-Stage Training Pipeline of \model{}}

We introduce a three-stage training pipeline designed to incrementally build the TTS model's feature control capability:

\begin{itemize}[leftmargin=12pt]
    \item \textbf{Stage 1: Pre-Training.} Establishes a foundational TTS capability by training the LLM to generate speech tokens from text.
    \item \textbf{Stage 2: Supervised Fine-Tuning~(SFT).} Teaches the LLM to generate speech conditioned on specific vocal features ($F_v$), enabling fine-grained control.
    \item \textbf{Stage 3: Preference Optimization~(PO).} Refines the model by preference optimizing, mitigating failure modes and enhancing the precision of feature control.
\end{itemize}

\paragraph{Stage 1: Pre-Training}
The objective of this stage is to equip the LLM with a fundamental text-to-speech capability, providing a robust weight initialization for subsequent stages. We use a large-scale corpus of speech-text pairs, $\mathcal{D}_{\text{pretrain}} = \{(x_i, S_i)\}$, where $x_i$ is the transcript and $S_i$ represents the corresponding discrete speech tokens. The model, denoted as policy $\pi_{\text{pre}}$, is trained using a standard causal language modeling objective to predict the next token autoregressively over the concatenated sequence of text and speech tokens. The training objective is:

$$
\mathcal{L}_{\text{Pre-Train}} = - \mathbb{E}_{(x, S) \sim \mathcal{D}_{\text{pretrain}}} \left[ \sum_{i=1}^{|x|+|S|} \log \pi_{\text{pre}}(y_i | y_{<i}) \right],
$$
where $Y = [x; S]$ is the concatenated sequence. This process trains the model on both text-to-text and text-to-speech-token generation, establishing a strong baseline.

\paragraph{Stage 2: Supervised Fine-Tuning}
The SFT stage aims to instill fine-grained controllability by training the model to generate speech conditioned on both the transcript and a set of explicit, verbalized vocal features. This process, illustrated in Figure~\ref{fig:framework}, trains the model to associate textual vocal features with  corresponding discrete speech tokens.

The process begins with a diverse corpus of speech-text pairs, $\mathcal{D}_{\text{raw}} = \{(A_i, x_i)\}$. For each pair, we perform word-level alignment and segment the speech. For each segment, we extract the its vocal features and verbalize them into a structured, human-readable textual representation, $F_v$ (e.g., a JSON-like string), which makes the vocal features directly controllable by a text-only LLM.

During this stage, we fine-tune the policy $\pi_{\text{sft}}$. The input sequence is formed by concatenating the transcript $x$, the verbalized features $F_v$, and the speech tokens $S$. The model is trained to predict the next token autoregressively by minimizing the cross-entropy loss:
$$
\mathcal{L}_{\text{SFT}} = - \mathbb{E}_{(x, F_v, S) \sim \mathcal{D}_{\text{sft}}} \left[ \sum_{i=1}^{|x|+|F_v|+|S|} \log \pi_{\text{sft}}(y_i | y_{<i}) \right],
$$
where $Y = [x; F_v; S]$ is the concatenated sequence. This objective teaches the model to generate speech tokens $S$ that adhere to the vocal plan specified by $F_v$.

\paragraph{Stage 3: Preference Optimazation}
Although SFT offers a direct mechanism for control, the resulting model, $\pi_{\text{sft}}$, is still prone to certain failure modes. These include a high Word Error Rate (WER), an unnaturally slow speaking rate and insufficient expressiveness. To overcome these limitations, we employ a subsequent preference optimazation stage. The central principle is to construct a preference dataset, $\mathcal{D}_{\text{pref}}$, designed to align the model's outputs with more desirable vocal features, crucially without the need for manually annotated expressive data.

The construction of $\mathcal{D}_{\text{pref}}$ involves the following steps:

\begin{itemize}[leftmargin=12pt]
    \item \textbf{Initial Generation and Rejection Sampling}: For each text prompt $x$ in a corpus $\mathcal{T}$, we use the pre-trained model $\pi_{\text{pre}}$ from Stage 1 to synthesize an speech sample $s_{\text{base}}$. Samples are designated as \emph{rejected} ($s_l$) if they exhibit a high WER or a slow speech rate (SR). The corresponding speech tokens $S_{\text{base}}$ are stored as the rejected sequence $S_l$.
    $$
    s_l \leftarrow s_{\text{base}} \quad \text{if} \quad \text{WER}(s_{\text{base}}, x) > \tau_{\text{wer\_high}} \quad \text{or} \quad \text{SR}(s_{\text{base}}) < \tau_{\text{sr}}.
    $$
    \item \textbf{Preferred Data Construction}: For each text $x$ corresponding to a rejected sample, we use our SFT model $\pi_{\text{sft}}$ to generate a new candidate speech $\hat{s}$. These candidates are accepted as \emph{chosen} samples ($s_w$) if they meet the quality criteria (low WER and adequate SR). The corresponding features and tokens $(F_{v,w}, S_w)$ are stored.
    $$
    s_w \leftarrow \hat{s} \quad \text{if} \quad \text{WER}(\hat{s}, x) \le \tau_{\text{wer\_high}} \quad \text{and} \quad \text{SR}(\hat{s}) \ge \tau_{\text{sr}}.
    $$
    \item \textbf{Preference Dataset Construction}: This filtering process yields pairs of chosen sequences $(F_{v,w}, S_w)$ and rejected sequences $S_l$. To create a controlled comparison, we form preference tuples where the model learns to prefer $S_w$ over $S_l$ under the same vocal plan, $F_{v,w}$. This setup creates a powerful learning signal: because $S_l$ was generated without knowledge of $F_{v,w}$, while $S_w$ was explicitly conditioned on it, teaching the model to prefer $S_w$ over $S_l$ not only improves general quality but also implicitly reinforces the model's ability to follow the specified vocal features. The final dataset consists of tuples: $\mathcal{D}_{\text{pref}} = \{(x_i, F_{v,w,i}, S_{w,i}, S_{l,i})\}$.

\end{itemize}

Finally, we fine-tune the model using Anchored Preference Optimization (APO-down)~\citep{d2025anchored}, with the SFT model serving as the reference policy ($\pi_{\text{ref}} = \pi_{\text{sft}}$). The APO-down objective penalizes deviations from the reference for the chosen sequence $S_w$ while maximizing the reward margin between the chosen ($S_w$) and rejected ($S_l$) sequences, given the shared prefix $(x, F_{v,w})$:
$$
\mathcal{L}_{\text{down}}^{\text{APO}}(\theta) = \mathbb{E}_{(x, F_{v,w}, S_w, S_l) \sim \mathcal{D}_{\text{pref}}} \left[ \underbrace{\sigma\!\left(r_{\theta}\!\left(x, F_{v,w}, S_{w}\right)\right)}_{\text{Term 1}} - \underbrace{\sigma\!\left(r_{\theta}\!\left(x, F_{v,w}, S_{w}\right) - r_{\theta}\!\left(x, F_{v,w}, S_{l}\right)\right)}_{\text{Term 2}} \right],
$$
where $r_{\theta}(x, F_v, S) = \beta \log \big(\pi_{\theta}(S \mid x, F_v) / \pi_{\text{ref}}(S \mid x, F_v)\big)$ is the implicit reward. Term 1 anchors the policy to the SFT model for chosen samples, while Term 2 maximizes the preference margin. This dual objective allows the model to mitigate common failure modes without requiring any explicitly labeled expressive data.

\section{Experiment}

\subsection{Experimental Setup}

\paragraph{Training \method{}}

The pre-training stage equips the LLM with the fundamental capability of converting text into a corresponding sequence of speech tokens, establishing a strong foundation for standard TTS before introducing complex instruction-following behavior. We use the VoxBox dataset \citep{wang2025spark}, a large-scale, multi-speaker English speech corpus of approximately 103K hours. The speech is tokenized into discrete vocal units using the official CosyVoice2 tokenizer. To maximize throughput, we pack tokenized sequences into 4096-token chunks, reducing padding overhead. Pre-training is conducted on 80 GPUs for 3 epochs (approximately one day), using AdamW with a learning rate of 1e-4, 500 warmup steps, a global batch size of 640, and DeepSpeed ZeRO-2 for memory optimization.

Our post-training process consists of SFT and PO. A key challenge in preparing the SFT data is that our speech decoder cannot perfectly reconstruct original speech from its quantized tokens. To ensure the vocal features are faithfully synthesizable, we derive them from speech that has been reconstructed by the decoder itself.

Our SFT dataset is compiled from two primary sources. First, we take a diverse collection of expressive speech corpora~\citep{veaux2017cstr,nagraniy2017voxceleb,chung2018voxceleb2,richter2024ears,nguyen2023expresso,yang2025emovoice,wang2025capspeech}, pass the speech through our decoder for reconstruction, and then extract features from the synthesized output. Second, we collect colloquial sentences from the Synthetic-Persona-Chat dataset~\citep{jandaghi-etal-2024-faithful} and synthesize them. We then apply a filtering process to the combined data, removing samples with a high Word Error Rate (WER), which indicates potential misalignments, or an abnormally slow speaking rate. $\tau_{\text{wer\_high}}$ is 0.1 and $\tau_{\text{sr}}$ is 1.5 words per seconds.This results in a final SFT dataset of 377,619 utterances, totaling over 500 hours (see Appendix for a detailed distribution). For the PO stage, we collected a dataset of 9,823 preference samples.

The feature extraction pipeline for this data begins with grounding features in semantically meaningful units. We first obtain word-level timestamps for each speech sample using a pre-trained model~\footnote{\url{https://huggingface.co/facebook/wav2vec2-large-960h-lv60-self}}. Since individual words are often too short to carry significant prosodic information, we merge adjacent words into segments until each segment's duration exceeds a one-second threshold, ensuring a stable and analyzable prosodic contour. Finally, we use the Parselmouth library~\footnote{\url{https://github.com/YannickJadoul/Parselmouth}} to extract a  set of vocal features from these segments. The model is trained with SFT for 3 epochs, followed by 1 epoch of APO-down.

\paragraph{Benchmarks and Evaluation.}
We selected two distinct benchmarks to rigorously test different facets of our model's performance:

\begin{itemize}[leftmargin=12pt]
    \item \textbf{TTS Intelligibility}: We use the test set from the \textbf{Seed-TTS} benchmark~\citep{anastassiou2024seed}, which is designed for assessing speech synthesis from short speech prompts. Performance is measured by \textbf{Word Error Rate (WER)}, calculated with pre-trained ASR models~\footnote{\url{https://huggingface.co/openai/whisper-large-v3}}. A lower WER score signifies higher intelligibility.

    \item \textbf{Emotion Control}: This is assessed on a curated test set from the \textbf{Emotion} dataset~\citep{saravia2018carer}. We use includes 100 samples for each of five emotions (joy, sadness, anger, surprise, and fear). We measure performance using \textbf{Emotion Classification Accuracy}. This metric is derived by employing Google's Gemini-2.5-Pro to classify the emotion of the synthesized speech. A higher accuracy indicates a greater success rate in generating perceptually accurate emotional speech. The prompt template for this evaluation is provided in the Appendix.
\end{itemize}

\subsection{Main Results}

\begin{table}[t]
\centering
\caption{Performance on the English TTS Benchmark. \method{} demonstrates superior emotion ability (Acc.) while maintaining high intelligibility (WER).}
\label{tab:tts_benchmark}
\begin{tabular}{l rrr cc}
\toprule
\multirow{2}{*}{\textbf{Model}}  &   \multirow{2}{*}{\textbf{Size}}  &   {\textbf{Pre-Train}}  &   {\textbf{Instruction}}   &   \textbf{Seed-TTS} & {\textbf{Emotion}}\\
\cmidrule(lr){5-5}\cmidrule(lr){6-6}
 &  &  \bf Data (Hours) & \bf Data (Hours)  &  \textbf{WER ($\downarrow$)} &  \textbf{Acc. ($\uparrow$)}\\
\midrule
\multicolumn{5}{l}{\bf Close-Source}\\
~~~Minimax-2.5-HD      & -& - & - & 1.5 & 48.6\\
~~~Minimax-2.5-Turbo   & -& - & - & 1.5 & 46.4\\
~~~Minimax-2.0-HD      & -& - & - & 1.5& 39.2 \\
\midrule
\multicolumn{5}{l}{\bf Open-Source}\\
~~~Spark-TTS       & 0.5B & 103K & 0     & 1.9 & 27.4\\
~~~CosyVoice       & 0.3B & 172K & 556   & 3.4 & 43.8\\
~~~CosyVoice2      & 0.5B & 167K & 1,500 & 2.1 &37.8\\
~~~Higgs speech V2  & 3.0B  & $>$10,000K  & -  & 1.8  & 23.5\\
\midrule
\multirow{2}{*}{\method{} (Ours)} &   0.5B    &   \multirow{2}{*}{103K} & \multirow{2}{*}{0}  &   2.9 &   52.8\\
                        &   1.7B    &   &  &   2.5 &   \bf57.6\\
\bottomrule
\end{tabular}
\end{table}

\paragraph{\method{} demonstrates strong emotion control performance while maintaining high intelligibility.}
As shown in Table~\ref{tab:tts_benchmark}, \method-1.7B achieves 57.6\% accuracy on the Emotion benchmark, surpassing the strongest closed-source baseline, Minimax-2.5-HD (48.6\%), by 9.0 absolute points. It also outperforms all open-source systems by a wide margin, e.g., +13.8 points over CosyVoice (43.8\%). On Seed-TTS, our 1.7B model attains a WER of 2.5 -- competitive with high-quality open models (better than CosyVoice at 3.4, slightly above CosyVoice2 at 2.1 and Spark-TTS at 1.9) and within a small gap of the closed-source Minimax series (1.5). These results validate that our decoupled ``conductor–orchestra” design substantially enhances emotional expressiveness without sacrificing intelligibility.

\paragraph{\method{} achieves substantial gains without manual instruction data.}
Our \method{} framework achieves these results with 0 hours of manually annotated instruction data, in contrast to CosyVoice and CosyVoice2, which use 556 and 1,500 hours respectively yet underperform on emotion accuracy (43.8\% and 37.8\%). Preference optimization over textual vocal plans yields consistent improvements over SFT alone: for the 1.7B model, emotion accuracy increases from 52.2\% (SFT) to 57.6\% (Instruct, +5.4 points) while WER improves from 2.9 to 2.5. Even at 0.5B, instruction tuning further boosts accuracy from 51.6\% to 52.8\% (+1.2). These gains directly confirm the effectiveness of \method{}.

\paragraph{\method{} demonstrates strong scalability with model size.}
Moving from 0.5B to 1.7B parameters improves emotion accuracy from 52.8\% to 57.6\% (+4.8) and reduces WER from 2.9 to 2.5 for the instruction-tuned models. This trend demonstrates the scalability of our method, and showcasing its consistent performance benefits across different model sizes..

\subsection{Human Evaluation of Instruction-Following}
\label{sec:human_evaluation}

To assess our model's performance on controllable TTS with free-form instructions, we create a specialized test set. We begin by sourcing 50 diverse social situations from the Social IQa benchmark~\citep{sap-etal-2019-social}, chosen for its rich contextual and emotional nuance. For each situation, we utilize \textit{Gemini 2.5 Pro} to generate a challenging test case. The model is prompted to produce two outputs: first, a detailed, role-playing style instruction framed in a second-person narrative, which specifies the desired persona and delivery style. Second, it generates a corresponding target utterance to be synthesized. This pipeline yield a high-quality benchmark of 50 pairs, specifically designed to test the model's ability to follow complex, descriptive instructions beyond simple labels.

\begin{wraptable}{r}{6cm}
\vspace{-10pt}
\centering
\caption{Human preference evaluation for instruction following TTS.}
\label{tab:human_eval}
\begin{tabular}{l r}
\toprule
\textbf{Compare with} &   \bf Win Rate\\
\midrule
CosyVoice & 56\%\\
Minimax-2.5-HD & 30\% \\
\bottomrule
\end{tabular}
\end{wraptable}

We found that long instructions were incorrectly synthesized by CosyVoice. To mitigate this, we use Gemini 2.5 Pro to map each detailed instruction to a discrete emotion label (Neutral + 6 Ekman emotions), which was then fed to CosyVoice. This label-based approach was also necessary for Minimax 2.5, which only accepts emotion labels as input.
We all the prompt template in the appendix.
We performed a human evaluation comparing \method{} with the top-performing open-source (CosyVoice) and closed-source (Minimax-2.5-HD) models. The evaluation involved three trained annotators with a Cohen's Kappa of 0.61. As shown in Table~\ref{tab:human_eval}, \method{} achieves performance comparable to CosyVoice but is outperformed by the commercial system Minimax-2.5-HD, falling short in aspects of fluency and naturalness.

\subsection{Cross-Lingual Generalization}
\label{sec:cross_lingual}

\begin{wraptable}{r}{6.6cm}
\vspace{-15pt}
\setlength{\tabcolsep}{3pt}
\centering
\caption{Performance on the {\bf Chinese} TTS Benchmark.  \method{} only uses English data for feature control training, yet demonstrates {\bf strong zero-shot generalization}.}
\label{tab:chinese_tts_benchmark}
\begin{tabular}{l cc}
\toprule
\multirow{2}{*}{\textbf{Model}}  &  \textbf{Seed-TTS} & {\textbf{Emotion}}\\
\cmidrule(lr){2-2}\cmidrule(lr){3-3}
 &  \textbf{WER ($\downarrow$)} &  \textbf{Acc. ($\uparrow$)}\\
\midrule
\multicolumn{3}{l}{\bf Close-Source}\\
~~~Minimax-2.5-HD      & 0.9& 49.0\\
~~~Minimax-2.5-Turbo   & 1.0& 50.6\\
~~~Minimax-2.0-HD      &0.9 & 48.8\\
\midrule
\multicolumn{3}{l}{\bf Open-Source}\\
~~~Spark-TTS       & 1.5& 29.2\\
~~~CosyVoice       & 2.1& 52.0\\
~~~CosyVoice2      & 2.0& 42.0\\
~~~Higgs speech V2  & 1.2    & 28.8\\
\midrule
\method-1.7B     & 2.1& \textbf{56.2}\\
\bottomrule
\end{tabular}
\vspace{-20pt}
\end{wraptable}

A significant and surprising finding is the model's ability to generalize to languages not seen during the \model{} post-training stage. We evaluated this by testing on a Chinese emotion benchmark, employing the same methodology as the English evaluation, with the text and instructions translated into Chinese by \textit{Gemini 2.5 Pro}. Notably, this cross-lingual generalization occurs despite the post-training stage being conducted exclusively on English data, demonstrating a strong zero-shot transfer capability.

\paragraph{\method{} demonstrates remarkable zero-shot cross-lingual generalization, applying feature control  ability to languages entirely unseen during post-training.}
As shown in Table~\ref{tab:chinese_tts_benchmark}, \model-1.7B achieves a 56.2\% accuracy on the Chinese emotion benchmark. This result is not only strong in absolute terms but also surpasses leading models that are either native to or heavily optimized for Chinese, such as CosyVoice (52.0\%) and the closed-source Minimax-2.5-Turbo (50.6\%). This performance is achieved without any Chinese instruction data, highlighting a key advantage of our ``operationalism'' paradigm.

\subsection{Component Analysis}
\label{sec:ablation}

We conduct a series of in-depth analyses to better understand the capabilities of \method. Otherwise stated, we report the results of \model-1.7B on the English Emotion benchmark.

\begin{figure}[h]
    \centering
    \subfloat[Impact of \model{} stages]{\includegraphics[width=0.3\linewidth]{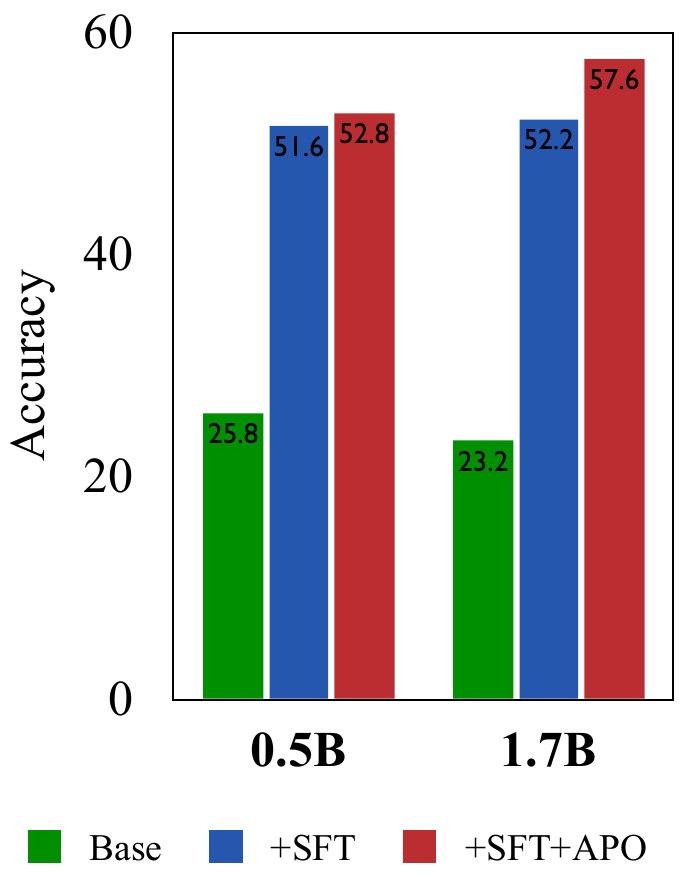}
    \label{fig:impact_stage}
    } \hspace{0.1\linewidth}
    \subfloat[Impact of LLMs]{\includegraphics[width=0.3\linewidth]{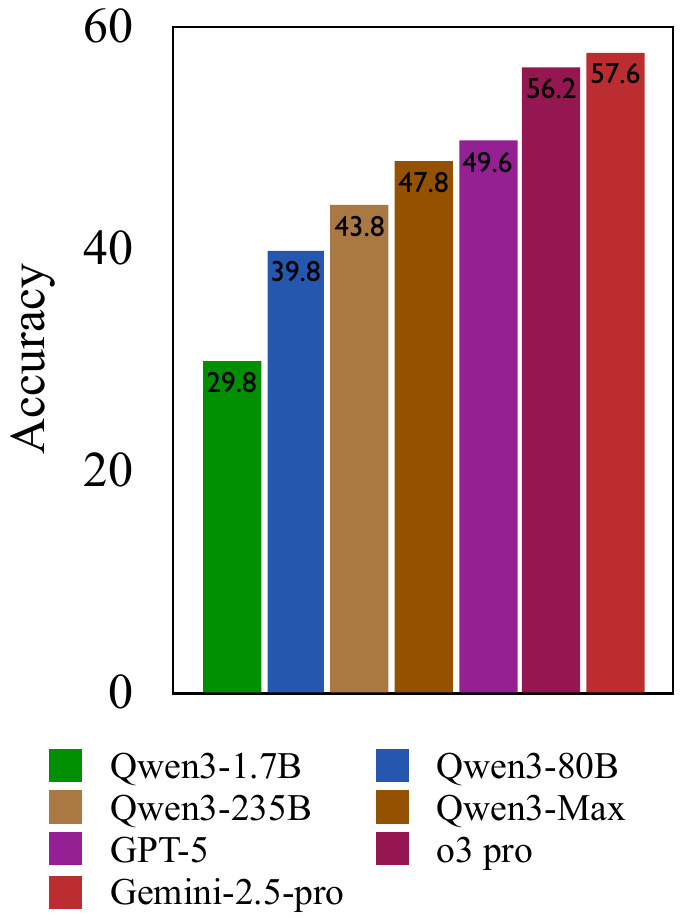}
    \label{fig:impact_feature_extractor}
    }
    \caption{Analysis of the key components of the proposed \method{}.}
    \label{fig:component}
\end{figure}

\paragraph{Each stage of the \model{} framework significantly contributes to emotional expressiveness.}
We perform a step-by-step ablation to examine the effectiveness of each stage in our proposed \model{} framework. As illustrated in Figure~\ref{fig:impact_stage}, the base model, trained only on foundational TTS without any instruction tuning, yields poor performance on the English Emotion benchmark  --  achieving just 23.2\% accuracy for the 1.7B model. Incorporating the SFT stage causes a dramatic improvement, boosting the accuracy to 52.2\% (+29.0 points), showing that teaching the model to generate and condition on verbalized vocal plans is key to enabling stylistic control. Adding the APO-based preference optimization further improves performance to 57.6\% (+5.4 over SFT), illustrating the importance of our post-training strategy. Consistent gains are observed for the smaller 0.5B model (25.8 $\rightarrow$ 51.6 $\rightarrow$ 52.8), demonstrating that the framework is effective across model scales. These results validate the design of \model{} in sequentially teaching foundational TTS capability  and improving control quality.

\paragraph{\model{} enables scalable leverage of LLM linguistic intelligence for emotional control.}
To demonstrate how our framework leverages the linguistic intelligence of Large Language Models (LLMs), we performed an experiment to measure the impact of the vocal feature generator on final synthesis quality. We used a fixed \method{} model and generated vocal plans at inference time using a range of LLMs with varying capabilities. As illustrated in Figure~\ref{fig:impact_feature_extractor}, the results show a clear, positive correlation between the performance of the LLM and the emotion accuracy of the synthesized speech. The accuracy climbs steadily from 29.8\% with Qwen3-1.7B to 57.6\% with Gemini-2.5-Pro, with intermediate models like Qwen3-80B (39.8\%) and Qwen3-Max (47.8\%) falling along this expected trajectory. 
These findings strongly support our core claim: representing speech as vocal features allows the synthesis model to directly benefit from advances in LLMs. This highlights a key advantage of our decoupled ``conductor–orchestra" design: its modularity. Even our compact 1.7B model can tap into the power of a much larger model like Gemini-2.5-Pro at inference time, effectively upgrading its expressive capability without any modification to the model.

\section{Related Work}

\paragraph{Controllable Speech Synthesis}
Controllable speech synthesis is typically classified into three primary paradigms. The first, style tagging, employs discrete labels (e.g., emotion, gender) to guide the synthesis process~\citep{guo2023prompttts,wang2025spark,zhang2025minimax}. While conceptually simple, this approach is restricted to a predefined set of styles, which fundamentally limits its expressive range. 
The second paradigm leverages reference speech to enable few-shot or zero-shot speaker adaptation. This is accomplished by extracting speaker embeddings from short speech samples and conditioning the TTS decoder on them -- a technique proven effective for voice cloning and style transfer~\citep{jiang2024megatts2boostingprompting,ji2025controlspeechsimultaneousindependentzeroshot,li2024stylettszsefficienthighqualityzeroshot}. 
The third and most flexible paradigm, instruction-guided control, conceptualizes TTS as a task of interpreting natural language instructions. Frameworks such as VoxInstruct exemplify this approach, guiding synthesis 
with free-form instructions~\citep{du2024cosyvoice,du2024cosyvoice2scalablestreaming}. 
However, these instruction-following methods are constrained by the high cost and difficulty of creating large-scale, annotated instruction-speech datasets, which limits their generalization and performance.

In contrast, our approach circumvents the need for manually annotated data by leveraging a powerful LLM. This enables robust, zero-shot generalization to unseen instructions, generating vocal features that exhibit high fidelity to the prompts while affording a high degree of control. Our method thus addresses the key limitations of data scarcity and annotation cost in instruction-guided TTS, showing significant promise for future research in expressive and controllable speech synthesis.


\paragraph{Multimodal Reasoning}
The remarkable reasoning capabilities of LLMs have catalyzed extensive research into extending these faculties to multimodal domains. Early efforts sought to enhance multimodal understanding by employing techniques such as reinforcement learning to better align visual and textual representations~\citep{hong2025glm,huang2025visionr1incentivizingreasoningcapability,luo2025guir1generalistr1style,shen2025vlmr1stablegeneralizabler1style}. More recent and prominent approaches aim for a deeper integration of reasoning. One prominent direction integrates multimodal information as intermediate steps within a reasoning chain, analogous to a ``chain of thought'', to derive conclusions~\citep{su2025openthinkimg,zheng2025deepeyes,zhang2025chain}. Another emerging strategy involves performing explicit, text-based reasoning prior to the final multimodal generation, thereby ensuring the output is logically grounded and coherent with the input prompt~\citep{liao2025imagegen,jiang2025t2i,huang2025interleavingreasoningbettertexttoimage}. While powerful, these methods typically rely on training large-scale, end-to-end multimodal models -- a process that is computationally intensive and demands vast quantities of aligned data.

In contrast to building new large-scale models, our work leverages existing text-only LLMs. We achieve this by representing multimodal information as quantifiable features that an LLM can manipulate based on user commands. This strategy is computationally efficient and scalable, as system performance advances with the underlying LLM without requiring retraining.

\section{Conclusion}

In this paper, we address a key limitation in current speech synthesis systems: the underutilization of the linguistic intelligence of LLMs. We introduce a new paradigm inspired by ``operationalism'', which decouples instruction understanding from speech generation by first translating instructions into  quantifiable, interpretable vocal features. Our framework, \textbf{\method{}}, embodies this principle by using LLMs to generate a vocal ``plan'', which is then fed into a TTS model. We train this model using a three-stage training pipeline that requires no manual instruction data. Our empirical results demonstrate the effectiveness of this approach. \method{} achieves strong performance in emotional speech synthesis and shows that its capabilities scale positively with the linguistic intelligence of LLMs. Furthermore, it exhibits powerful zero-shot cross-lingual generalization. 

The central claim of this work is that the most effective path to leveraging the intelligence of LLMs lies in the textual representation of other modalities.. This principle delineates a novel and promising direction for MLLM research. The prospective applications of this operationalist approach are extensive, which can be extended to other modalities, such as video and music. Furthermore, within the speech domain, further investigation should focus on enriching vocal plans to capture finer-grained paralinguistic features, including emphatic stress, and non-verbal vocalizations.

\bibliography{ref}

\begin{thebibliography}{35}
\providecommand{\natexlab}[1]{#1}
\providecommand{\url}[1]{\texttt{#1}}
\expandafter\ifx\csname urlstyle\endcsname\relax
  \providecommand{\doi}[1]{doi: #1}\else
  \providecommand{\doi}{doi: \begingroup \urlstyle{rm}\Url}\fi

\bibitem[Anastassiou et~al.(2024)Anastassiou, Chen, Chen, Chen, Chen, Chen, Cong, Deng, Ding, Gao, et~al.]{anastassiou2024seed}
Philip Anastassiou, Jiawei Chen, Jitong Chen, Yuanzhe Chen, Zhuo Chen, Ziyi Chen, Jian Cong, Lelai Deng, Chuang Ding, Lu~Gao, et~al.
\newblock Seed-tts: A family of high-quality versatile speech generation models.
\newblock \emph{arXiv preprint arXiv:2406.02430}, 2024.

\bibitem[Chung et~al.(2018)Chung, Nagrani, and Zisserman]{chung2018voxceleb2}
Joon~Son Chung, Arsha Nagrani, and Andrew Zisserman.
\newblock Voxceleb2: Deep speaker recognition.
\newblock In \emph{Proceedings of the Annual Conference of the International Speech Communication Association, INTERSPEECH}, volume 2018, pp.\  1086--1090, 2018.

\bibitem[Deng et~al.(2025)Deng, Zhu, Li, Gou, Li, Wang, Zhong, Yu, Nie, Song, et~al.]{deng2025emerging}
Chaorui Deng, Deyao Zhu, Kunchang Li, Chenhui Gou, Feng Li, Zeyu Wang, Shu Zhong, Weihao Yu, Xiaonan Nie, Ziang Song, et~al.
\newblock Emerging properties in unified multimodal pretraining.
\newblock \emph{arXiv preprint arXiv:2505.14683}, 2025.

\bibitem[D'Oosterlinck et~al.(2025)D'Oosterlinck, Xu, Develder, Demeester, Singh, Potts, Kiela, and Mehri]{d2025anchored}
Karel D'Oosterlinck, Winnie Xu, Chris Develder, Thomas Demeester, Amanpreet Singh, Christopher Potts, Douwe Kiela, and Shikib Mehri.
\newblock Anchored preference optimization and contrastive revisions: Addressing underspecification in alignment.
\newblock \emph{Transactions of the Association for Computational Linguistics}, 13:\penalty0 442--460, 2025.

\bibitem[Du et~al.(2024{\natexlab{a}})Du, Wang, Chen, Shi, Lv, Zhao, Gao, Yang, Gao, Wang, Yu, Liu, Sheng, Gu, Deng, Wang, Zhang, Yan, and Zhou]{du2024cosyvoice2scalablestreaming}
Zhihao Du, Yuxuan Wang, Qian Chen, Xian Shi, Xiang Lv, Tianyu Zhao, Zhifu Gao, Yexin Yang, Changfeng Gao, Hui Wang, Fan Yu, Huadai Liu, Zhengyan Sheng, Yue Gu, Chong Deng, Wen Wang, Shiliang Zhang, Zhijie Yan, and Jingren Zhou.
\newblock Cosyvoice 2: Scalable streaming speech synthesis with large language models, 2024{\natexlab{a}}.
\newblock URL \url{https://arxiv.org/abs/2412.10117}.

\bibitem[Du et~al.(2024{\natexlab{b}})Du, Wang, Chen, Shi, Lv, Zhao, Gao, Yang, Gao, Wang, et~al.]{du2024cosyvoice}
Zhihao Du, Yuxuan Wang, Qian Chen, Xian Shi, Xiang Lv, Tianyu Zhao, Zhifu Gao, Yexin Yang, Changfeng Gao, Hui Wang, et~al.
\newblock Cosyvoice 2: Scalable streaming speech synthesis with large language models.
\newblock \emph{arXiv preprint arXiv:2412.10117}, 2024{\natexlab{b}}.

\bibitem[Guo et~al.(2023)Guo, Leng, Wu, Zhao, and Tan]{guo2023prompttts}
Zhifang Guo, Yichong Leng, Yihan Wu, Sheng Zhao, and Xu~Tan.
\newblock Prompttts: Controllable text-to-speech with text descriptions.
\newblock In \emph{ICASSP 2023-2023 IEEE International Conference on Acoustics, Speech and Signal Processing (ICASSP)}, pp.\  1--5. IEEE, 2023.

\bibitem[Hong et~al.(2025)Hong, Yu, Gu, Wang, Gan, Tang, Cheng, Qi, Ji, Pan, et~al.]{hong2025glm}
Wenyi Hong, Wenmeng Yu, Xiaotao Gu, Guo Wang, Guobing Gan, Haomiao Tang, Jiale Cheng, Ji~Qi, Junhui Ji, Lihang Pan, et~al.
\newblock Glm-4.1 v-thinking: Towards versatile multimodal reasoning with scalable reinforcement learning.
\newblock \emph{arXiv e-prints}, pp.\  arXiv--2507, 2025.

\bibitem[Huang et~al.(2025{\natexlab{a}})Huang, Chen, Xie, Cao, Tang, Shen, Yin, Hu, Wang, Tang, Qiao, Guo, Hu, Yin, Torr, Cheng, Ouyang, and Lin]{huang2025interleavingreasoningbettertexttoimage}
Wenxuan Huang, Shuang Chen, Zheyong Xie, Shaosheng Cao, Shixiang Tang, Yufan Shen, Qingyu Yin, Wenbo Hu, Xiaoman Wang, Yuntian Tang, Junbo Qiao, Yue Guo, Yao Hu, Zhenfei Yin, Philip Torr, Yu~Cheng, Wanli Ouyang, and Shaohui Lin.
\newblock Interleaving reasoning for better text-to-image generation, 2025{\natexlab{a}}.
\newblock URL \url{https://arxiv.org/abs/2509.06945}.

\bibitem[Huang et~al.(2025{\natexlab{b}})Huang, Jia, Zhai, Cao, Ye, Zhao, Xu, Hu, and Lin]{huang2025visionr1incentivizingreasoningcapability}
Wenxuan Huang, Bohan Jia, Zijie Zhai, Shaosheng Cao, Zheyu Ye, Fei Zhao, Zhe Xu, Yao Hu, and Shaohui Lin.
\newblock Vision-r1: Incentivizing reasoning capability in multimodal large language models, 2025{\natexlab{b}}.
\newblock URL \url{https://arxiv.org/abs/2503.06749}.

\bibitem[Jandaghi et~al.(2024)Jandaghi, Sheng, Bai, Pujara, and Sidahmed]{jandaghi-etal-2024-faithful}
Pegah Jandaghi, Xianghai Sheng, Xinyi Bai, Jay Pujara, and Hakim Sidahmed.
\newblock Faithful persona-based conversational dataset generation with large language models.
\newblock In Elnaz Nouri, Abhinav Rastogi, Georgios Spithourakis, Bing Liu, Yun-Nung Chen, Yu~Li, Alon Albalak, Hiromi Wakaki, and Alexandros Papangelis (eds.), \emph{Proceedings of the 6th Workshop on NLP for Conversational AI (NLP4ConvAI 2024)}, pp.\  114--139, Bangkok, Thailand, August 2024. Association for Computational Linguistics.
\newblock URL \url{https://aclanthology.org/2024.nlp4convai-1.8/}.

\bibitem[Ji et~al.(2025)Ji, Chen, Wang, Zuo, Fang, Jiang, Huang, Wang, Cheng, Zheng, and Zhao]{ji2025controlspeechsimultaneousindependentzeroshot}
Shengpeng Ji, Qian Chen, Wen Wang, Jialong Zuo, Minghui Fang, Ziyue Jiang, Hai Huang, Zehan Wang, Xize Cheng, Siqi Zheng, and Zhou Zhao.
\newblock Controlspeech: Towards simultaneous and independent zero-shot speaker cloning and zero-shot language style control, 2025.
\newblock URL \url{https://arxiv.org/abs/2406.01205}.

\bibitem[Jiang et~al.(2025)Jiang, Guo, Zhang, Zong, Li, Zhuo, Yan, Heng, and Li]{jiang2025t2i}
Dongzhi Jiang, Ziyu Guo, Renrui Zhang, Zhuofan Zong, Hao Li, Le~Zhuo, Shilin Yan, Pheng-Ann Heng, and Hongsheng Li.
\newblock T2i-r1: Reinforcing image generation with collaborative semantic-level and token-level cot.
\newblock \emph{arXiv preprint arXiv:2505.00703}, 2025.

\bibitem[Jiang et~al.(2024)Jiang, Liu, Ren, He, Ye, Ji, Yang, Zhang, Wei, Wang, Yin, Ma, and Zhao]{jiang2024megatts2boostingprompting}
Ziyue Jiang, Jinglin Liu, Yi~Ren, Jinzheng He, Zhenhui Ye, Shengpeng Ji, Qian Yang, Chen Zhang, Pengfei Wei, Chunfeng Wang, Xiang Yin, Zejun Ma, and Zhou Zhao.
\newblock Mega-tts 2: Boosting prompting mechanisms for zero-shot speech synthesis, 2024.
\newblock URL \url{https://arxiv.org/abs/2307.07218}.

\bibitem[Li et~al.(2024)Li, Jiang, Han, and Mesgarani]{li2024stylettszsefficienthighqualityzeroshot}
Yinghao~Aaron Li, Xilin Jiang, Cong Han, and Nima Mesgarani.
\newblock Styletts-zs: Efficient high-quality zero-shot text-to-speech synthesis with distilled time-varying style diffusion, 2024.
\newblock URL \url{https://arxiv.org/abs/2409.10058}.

\bibitem[Liao et~al.(2025)Liao, Yang, Li, Li, Lin, Cheng, and Wang]{liao2025imagegen}
Jiaqi Liao, Zhengyuan Yang, Linjie Li, Dianqi Li, Kevin Lin, Yu~Cheng, and Lijuan Wang.
\newblock Imagegen-cot: Enhancing text-to-image in-context learning with chain-of-thought reasoning.
\newblock \emph{arXiv preprint arXiv:2503.19312}, 2025.

\bibitem[Livingstone \& Russo(2018)Livingstone and Russo]{livingstone2018ryerson}
Steven~R Livingstone and Frank~A Russo.
\newblock The ryerson audio-visual database of emotional speech and song (ravdess): A dynamic, multimodal set of facial and vocal expressions in north american english.
\newblock \emph{PloS one}, 13\penalty0 (5):\penalty0 e0196391, 2018.

\bibitem[Luo et~al.(2025)Luo, Wang, He, and Xia]{luo2025guir1generalistr1style}
Run Luo, Lu~Wang, Wanwei He, and Xiaobo Xia.
\newblock Gui-r1 : A generalist r1-style vision-language action model for gui agents, 2025.
\newblock URL \url{https://arxiv.org/abs/2504.10458}.

\bibitem[Nagraniy et~al.(2017)Nagraniy, Chungy, and Zisserman]{nagraniy2017voxceleb}
Arsha Nagraniy, Joon~Son Chungy, and Andrew Zisserman.
\newblock Voxceleb: A large-scale speaker identification dataset.
\newblock In \emph{Proceedings of the Annual Conference of the International Speech Communication Association, INTERSPEECH}, volume 2017, pp.\  2616--2620, 2017.

\bibitem[Nguyen et~al.(2023)Nguyen, Hsu, D'Avirro, Shi, Gat, Fazel-Zarandi, Remez, Copet, Synnaeve, Hassid, et~al.]{nguyen2023expresso}
Tu~Anh Nguyen, Wei-Ning Hsu, Antony D'Avirro, Bowen Shi, Itai Gat, Maryam Fazel-Zarandi, Tal Remez, Jade Copet, Gabriel Synnaeve, Michael Hassid, et~al.
\newblock Expresso: A benchmark and analysis of discrete expressive speech resynthesis.
\newblock In \emph{INTERSPEECH}, 2023.

\bibitem[Richter et~al.(2024)Richter, Wu, Krenn, Welker, Lay, Watanabe, Richard, and Gerkmann]{richter2024ears}
Julius Richter, Yi-Chiao Wu, Steven Krenn, Simon Welker, Bunlong Lay, Shinji Watanabe, Alexander Richard, and Timo Gerkmann.
\newblock Ears: An anechoic fullband speech dataset benchmarked for speech enhancement and dereverberation.
\newblock In \emph{Proc. Interspeech 2024}, pp.\  4873--4877, 2024.

\bibitem[Sap et~al.(2019)Sap, Rashkin, Chen, Le~Bras, and Choi]{sap-etal-2019-social}
Maarten Sap, Hannah Rashkin, Derek Chen, Ronan Le~Bras, and Yejin Choi.
\newblock Social {IQ}a: Commonsense reasoning about social interactions.
\newblock In Kentaro Inui, Jing Jiang, Vincent Ng, and Xiaojun Wan (eds.), \emph{Proceedings of the 2019 Conference on Empirical Methods in Natural Language Processing and the 9th International Joint Conference on Natural Language Processing (EMNLP-IJCNLP)}, pp.\  4463--4473, Hong Kong, China, November 2019. Association for Computational Linguistics.
\newblock \doi{10.18653/v1/D19-1454}.
\newblock URL \url{https://aclanthology.org/D19-1454/}.

\bibitem[Saravia et~al.(2018)Saravia, Liu, Huang, Wu, and Chen]{saravia2018carer}
Elvis Saravia, Hsien-Chi~Toby Liu, Yen-Hao Huang, Junlin Wu, and Yi-Shin Chen.
\newblock Carer: Contextualized affect representations for emotion recognition.
\newblock In \emph{Proceedings of the 2018 conference on empirical methods in natural language processing}, pp.\  3687--3697, 2018.

\bibitem[Shen et~al.(2025)Shen, Liu, Li, Fang, Ma, Liao, Shen, Zhang, Zhao, Zhang, Xu, and Zhao]{shen2025vlmr1stablegeneralizabler1style}
Haozhan Shen, Peng Liu, Jingcheng Li, Chunxin Fang, Yibo Ma, Jiajia Liao, Qiaoli Shen, Zilun Zhang, Kangjia Zhao, Qianqian Zhang, Ruochen Xu, and Tiancheng Zhao.
\newblock Vlm-r1: A stable and generalizable r1-style large vision-language model, 2025.
\newblock URL \url{https://arxiv.org/abs/2504.07615}.

\bibitem[Su et~al.(2025)Su, Li, Song, Hao, Yang, Zhang, Chen, Gu, Li, Qu, et~al.]{su2025openthinkimg}
Zhaochen Su, Linjie Li, Mingyang Song, Yunzhuo Hao, Zhengyuan Yang, Jun Zhang, Guanjie Chen, Jiawei Gu, Juntao Li, Xiaoye Qu, et~al.
\newblock Openthinkimg: Learning to think with images via visual tool reinforcement learning.
\newblock \emph{arXiv preprint arXiv:2505.08617}, 2025.

\bibitem[Veaux et~al.(2017)Veaux, Yamagishi, MacDonald, et~al.]{veaux2017cstr}
Christophe Veaux, Junichi Yamagishi, Kirsten MacDonald, et~al.
\newblock Cstr vctk corpus: English multi-speaker corpus for cstr voice cloning toolkit.
\newblock \emph{University of Edinburgh. The Centre for Speech Technology Research (CSTR)}, 6:\penalty0 15, 2017.

\bibitem[Wang et~al.(2023)Wang, Chen, Wu, Zhang, Zhou, Liu, Chen, Liu, Wang, Li, He, Zhao, and Wei]{wang2023neuralcodeclanguagemodels}
Chengyi Wang, Sanyuan Chen, Yu~Wu, Ziqiang Zhang, Long Zhou, Shujie Liu, Zhuo Chen, Yanqing Liu, Huaming Wang, Jinyu Li, Lei He, Sheng Zhao, and Furu Wei.
\newblock Neural codec language models are zero-shot text to speech synthesizers, 2023.
\newblock URL \url{https://arxiv.org/abs/2301.02111}.

\bibitem[Wang et~al.(2025{\natexlab{a}})Wang, Hai, Chong, Thakkar, Feng, Yang, Lee, Velazquez, Villalba, Qin, et~al.]{wang2025capspeech}
Helin Wang, Jiarui Hai, Dading Chong, Karan Thakkar, Tiantian Feng, Dongchao Yang, Junhyeok Lee, Laureano~Moro Velazquez, Jesus Villalba, Zengyi Qin, et~al.
\newblock Capspeech: Enabling downstream applications in style-captioned text-to-speech.
\newblock \emph{arXiv preprint arXiv:2506.02863}, 2025{\natexlab{a}}.

\bibitem[Wang et~al.(2025{\natexlab{b}})Wang, Jiang, Ma, Zhang, Liu, Li, Liang, Zheng, Wang, Feng, et~al.]{wang2025spark}
Xinsheng Wang, Mingqi Jiang, Ziyang Ma, Ziyu Zhang, Songxiang Liu, Linqin Li, Zheng Liang, Qixi Zheng, Rui Wang, Xiaoqin Feng, et~al.
\newblock Spark-tts: An efficient llm-based text-to-speech model with single-stream decoupled speech tokens.
\newblock \emph{arXiv preprint arXiv:2503.01710}, 2025{\natexlab{b}}.

\bibitem[Xu et~al.(2025)Xu, Guo, Hu, Chu, Wang, He, Wang, Shi, He, Zhu, et~al.]{xu2025qwen3}
Jin Xu, Zhifang Guo, Hangrui Hu, Yunfei Chu, Xiong Wang, Jinzheng He, Yuxuan Wang, Xian Shi, Ting He, Xinfa Zhu, et~al.
\newblock Qwen3-omni technical report.
\newblock \emph{arXiv preprint arXiv:2509.17765}, 2025.

\bibitem[Yang et~al.(2025)Yang, Yang, Chen, Ma, Chen, Wang, Wang, Yang, Niu, Liu, et~al.]{yang2025emovoice}
Guanrou Yang, Chen Yang, Qian Chen, Ziyang Ma, Wenxi Chen, Wen Wang, Tianrui Wang, Yifan Yang, Zhikang Niu, Wenrui Liu, et~al.
\newblock Emovoice: Llm-based emotional text-to-speech model with freestyle text prompting.
\newblock \emph{arXiv preprint arXiv:2504.12867}, 2025.

\bibitem[Zeng et~al.(2024)Zeng, Du, Liu, Wang, Jiang, Zhao, Dong, and Tang]{zeng2024glm}
Aohan Zeng, Zhengxiao Du, Mingdao Liu, Kedong Wang, Shengmin Jiang, Lei Zhao, Yuxiao Dong, and Jie Tang.
\newblock Glm-4-voice: Towards intelligent and human-like end-to-end spoken chatbot.
\newblock \emph{arXiv preprint arXiv:2412.02612}, 2024.

\bibitem[Zhang et~al.(2025{\natexlab{a}})Zhang, Guo, Yang, Yu, Zhang, Lei, Mai, Yan, Yang, Yang, et~al.]{zhang2025minimax}
Bowen Zhang, Congchao Guo, Geng Yang, Hang Yu, Haozhe Zhang, Heidi Lei, Jialong Mai, Junjie Yan, Kaiyue Yang, Mingqi Yang, et~al.
\newblock Minimax-speech: Intrinsic zero-shot text-to-speech with a learnable speaker encoder.
\newblock \emph{arXiv preprint arXiv:2505.07916}, 2025{\natexlab{a}}.

\bibitem[Zhang et~al.(2025{\natexlab{b}})Zhang, Gao, Zhang, Li, Zhang, Liu, Yuan, Wu, Jia, Zhu, et~al.]{zhang2025chain}
Xintong Zhang, Zhi Gao, Bofei Zhang, Pengxiang Li, Xiaowen Zhang, Yang Liu, Tao Yuan, Yuwei Wu, Yunde Jia, Song-Chun Zhu, et~al.
\newblock Chain-of-focus: Adaptive visual search and zooming for multimodal reasoning via rl.
\newblock \emph{arXiv preprint arXiv:2505.15436}, 2025{\natexlab{b}}.

\bibitem[Zheng et~al.(2025)Zheng, Yang, Hong, Zhao, Xu, Yang, Shen, and Yu]{zheng2025deepeyes}
Ziwei Zheng, Michael Yang, Jack Hong, Chenxiao Zhao, Guohai Xu, Le~Yang, Chao Shen, and Xing Yu.
\newblock Deepeyes: Incentivizing" thinking with images" via reinforcement learning.
\newblock \emph{arXiv preprint arXiv:2505.14362}, 2025.

\end{thebibliography}


\begin{thebibliography}{44}
\providecommand{\natexlab}[1]{#1}
\providecommand{\url}[1]{\texttt{#1}}
\expandafter\ifx\csname urlstyle\endcsname\relax
  \providecommand{\doi}[1]{doi: #1}\else
  \providecommand{\doi}{doi: \begingroup \urlstyle{rm}\Url}\fi

\bibitem[qwe(2024)]{qwen2}
Qwen2 technical report.
\newblock 2024.

\bibitem[Abdin et~al.(2024)Abdin, Jacobs, Awan, Aneja, Awadallah, Awadalla, Bach, Bahree, Bakhtiari, Behl, et~al.]{abdin2024phi}
Marah Abdin, Sam~Ade Jacobs, Ammar~Ahmad Awan, Jyoti Aneja, Ahmed Awadallah, Hany Awadalla, Nguyen Bach, Amit Bahree, Arash Bakhtiari, Harkirat Behl, et~al.
\newblock Phi-3 technical report: A highly capable language model locally on your phone.
\newblock \emph{arXiv preprint arXiv:2404.14219}, 2024.

\bibitem[Achiam et~al.(2023)Achiam, Adler, Agarwal, Ahmad, Akkaya, Aleman, Almeida, Altenschmidt, Altman, Anadkat, et~al.]{achiam2023gpt}
Josh Achiam, Steven Adler, Sandhini Agarwal, Lama Ahmad, Ilge Akkaya, Florencia~Leoni Aleman, Diogo Almeida, Janko Altenschmidt, Sam Altman, Shyamal Anadkat, et~al.
\newblock Gpt-4 technical report.
\newblock \emph{arXiv preprint arXiv:2303.08774}, 2023.

\bibitem[Bai et~al.(2024)Bai, Du, Liang, Jin, Liu, Zhou, Zheng, Zhang, Ma, Wang, et~al.]{bai2024coig}
Yuelin Bai, Xinrun Du, Yiming Liang, Yonggang Jin, Ziqiang Liu, Junting Zhou, Tianyu Zheng, Xincheng Zhang, Nuo Ma, Zekun Wang, et~al.
\newblock Coig-cqia: Quality is all you need for chinese instruction fine-tuning.
\newblock \emph{arXiv preprint arXiv:2403.18058}, 2024.

\bibitem[Bauer et~al.(2024)Bauer, Trapp, Stenger, Leppich, Kounev, Leznik, Chard, and Foster]{bauer2024comprehensive}
Andr{\'e} Bauer, Simon Trapp, Michael Stenger, Robert Leppich, Samuel Kounev, Mark Leznik, Kyle Chard, and Ian Foster.
\newblock Comprehensive exploration of synthetic data generation: A survey.
\newblock \emph{arXiv preprint arXiv:2401.02524}, 2024.

\bibitem[Bi et~al.(2024)Bi, Chen, Chen, Chen, Dai, Deng, Ding, Dong, Du, Fu, et~al.]{bi2024deepseek}
Xiao Bi, Deli Chen, Guanting Chen, Shanhuang Chen, Damai Dai, Chengqi Deng, Honghui Ding, Kai Dong, Qiushi Du, Zhe Fu, et~al.
\newblock Deepseek llm: Scaling open-source language models with longtermism.
\newblock \emph{arXiv preprint arXiv:2401.02954}, 2024.

\bibitem[Broder(1997)]{broder1997resemblance}
Andrei~Z Broder.
\newblock On the resemblance and containment of documents.
\newblock In \emph{Proceedings. Compression and Complexity of SEQUENCES 1997 (Cat. No. 97TB100171)}, pp.\  21--29. IEEE, 1997.

\bibitem[Cai et~al.(2023)Cai, Wang, Ma, Chen, and Zhou]{cai2023large}
Tianle Cai, Xuezhi Wang, Tengyu Ma, Xinyun Chen, and Denny Zhou.
\newblock Large language models as tool makers.
\newblock \emph{arXiv preprint arXiv:2305.17126}, 2023.

\bibitem[Chakraborty et~al.(2023)Chakraborty, Bedi, Zhu, An, Manocha, and Huang]{chakraborty2023possibilities}
Souradip Chakraborty, Amrit~Singh Bedi, Sicheng Zhu, Bang An, Dinesh Manocha, and Furong Huang.
\newblock On the possibilities of ai-generated text detection.
\newblock \emph{arXiv preprint arXiv:2304.04736}, 2023.

\bibitem[Chen et~al.(2022)Chen, Ma, Wang, and Cohen]{chen2022program}
Wenhu Chen, Xueguang Ma, Xinyi Wang, and William~W Cohen.
\newblock Program of thoughts prompting: Disentangling computation from reasoning for numerical reasoning tasks.
\newblock \emph{arXiv preprint arXiv:2211.12588}, 2022.

\bibitem[Choi \& Li(2024)Choi and Li]{choipicle}
Hyeong~Kyu Choi and Yixuan Li.
\newblock Picle: Eliciting diverse behaviors from large language models with persona in-context learning.
\newblock In \emph{Forty-first International Conference on Machine Learning}, 2024.

\bibitem[Computer(2023)]{together2023redpajama}
Together Computer.
\newblock Redpajama: an open dataset for training large language models, 2023.
\newblock URL \url{https://github.com/togethercomputer/RedPajama-Data}.

\bibitem[Del{\'e}tang et~al.(2023)Del{\'e}tang, Ruoss, Duquenne, Catt, Genewein, Mattern, Grau-Moya, Wenliang, Aitchison, Orseau, et~al.]{deletang2023language}
Gr{\'e}goire Del{\'e}tang, Anian Ruoss, Paul-Ambroise Duquenne, Elliot Catt, Tim Genewein, Christopher Mattern, Jordi Grau-Moya, Li~Kevin Wenliang, Matthew Aitchison, Laurent Orseau, et~al.
\newblock Language modeling is compression.
\newblock \emph{arXiv preprint arXiv:2309.10668}, 2023.

\bibitem[Dohmatob et~al.(2024)Dohmatob, Feng, Yang, Charton, and Kempe]{dohmatob2024tale}
Elvis Dohmatob, Yunzhen Feng, Pu~Yang, Francois Charton, and Julia Kempe.
\newblock A tale of tails: Model collapse as a change of scaling laws.
\newblock \emph{arXiv preprint arXiv:2402.07043}, 2024.

\bibitem[Gandhi et~al.(2023)Gandhi, Sadigh, and Goodman]{gandhi2023strategic}
Kanishk Gandhi, Dorsa Sadigh, and Noah~D Goodman.
\newblock Strategic reasoning with language models.
\newblock \emph{arXiv preprint arXiv:2305.19165}, 2023.

\bibitem[Ge et~al.(2024)Ge, Jing, Wang, Wang, Chen, and Wei]{ge2024incontext}
Tao Ge, Hu~Jing, Lei Wang, Xun Wang, Si-Qing Chen, and Furu Wei.
\newblock In-context autoencoder for context compression in a large language model.
\newblock In \emph{The Twelfth International Conference on Learning Representations}, 2024.
\newblock URL \url{https://openreview.net/forum?id=uREj4ZuGJE}.

\bibitem[Hendrycks et~al.(2021)Hendrycks, Burns, Kadavath, Arora, Basart, Tang, Song, and Steinhardt]{hendrycks2021measuring}
Dan Hendrycks, Collin Burns, Saurav Kadavath, Akul Arora, Steven Basart, Eric Tang, Dawn Song, and Jacob Steinhardt.
\newblock Measuring mathematical problem solving with the math dataset.
\newblock \emph{arXiv preprint arXiv:2103.03874}, 2021.

\bibitem[Huang et~al.(2024)Huang, Liu, Gong, Gou, Shen, Duan, and Chen]{huang2024key}
Yiming Huang, Xiao Liu, Yeyun Gong, Zhibin Gou, Yelong Shen, Nan Duan, and Weizhu Chen.
\newblock Key-point-driven data synthesis with its enhancement on mathematical reasoning.
\newblock \emph{arXiv preprint arXiv:2403.02333}, 2024.

\bibitem[Jandaghi et~al.(2023)Jandaghi, Sheng, Bai, Pujara, and Sidahmed]{jandaghi2023faithful}
Pegah Jandaghi, XiangHai Sheng, Xinyi Bai, Jay Pujara, and Hakim Sidahmed.
\newblock Faithful persona-based conversational dataset generation with large language models.
\newblock \emph{arXiv preprint arXiv:2312.10007}, 2023.

\bibitem[Kaplan et~al.(2020)Kaplan, McCandlish, Henighan, Brown, Chess, Child, Gray, Radford, Wu, and Amodei]{kaplan2020scaling}
Jared Kaplan, Sam McCandlish, Tom Henighan, Tom~B Brown, Benjamin Chess, Rewon Child, Scott Gray, Alec Radford, Jeffrey Wu, and Dario Amodei.
\newblock Scaling laws for neural language models.
\newblock \emph{arXiv preprint arXiv:2001.08361}, 2020.

\bibitem[Li et~al.(2024{\natexlab{a}})Li, Wang, Hu, Wei, Zheng, Hu, Zhang, and Peng]{li2024common}
Chen Li, Weiqi Wang, Jingcheng Hu, Yixuan Wei, Nanning Zheng, Han Hu, Zheng Zhang, and Houwen Peng.
\newblock Common 7b language models already possess strong math capabilities.
\newblock \emph{arXiv preprint arXiv:2403.04706}, 2024{\natexlab{a}}.

\bibitem[Li et~al.(2024{\natexlab{b}})Li, Dong, Tang, Wang, Zhang, Huang, Huang, Huang, Huang, Zhang, et~al.]{li2024synthetic}
Haoran Li, Qingxiu Dong, Zhengyang Tang, Chaojun Wang, Xingxing Zhang, Haoyang Huang, Shaohan Huang, Xiaolong Huang, Zeqiang Huang, Dongdong Zhang, et~al.
\newblock Synthetic data (almost) from scratch: Generalized instruction tuning for language models.
\newblock \emph{arXiv preprint arXiv:2402.13064}, 2024{\natexlab{b}}.

\bibitem[Li et~al.(2023{\natexlab{a}})Li, Mehrabi, Peris, Goyal, Chang, Galstyan, Zemel, and Gupta]{li2023steerability}
Junyi Li, Ninareh Mehrabi, Charith Peris, Palash Goyal, Kai-Wei Chang, Aram Galstyan, Richard Zemel, and Rahul Gupta.
\newblock On the steerability of large language models toward data-driven personas.
\newblock \emph{arXiv preprint arXiv:2311.04978}, 2023{\natexlab{a}}.

\bibitem[Li et~al.(2023{\natexlab{b}})Li, Bubeck, Eldan, Del~Giorno, Gunasekar, and Lee]{li2023textbooks}
Yuanzhi Li, S{\'e}bastien Bubeck, Ronen Eldan, Allie Del~Giorno, Suriya Gunasekar, and Yin~Tat Lee.
\newblock Textbooks are all you need ii: phi-1.5 technical report.
\newblock \emph{arXiv preprint arXiv:2309.05463}, 2023{\natexlab{b}}.

\bibitem[Liu et~al.(2024)Liu, Wei, Liu, Si, Zhang, Rao, Zheng, Peng, Yang, Zhou, et~al.]{liu2024best}
Ruibo Liu, Jerry Wei, Fangyu Liu, Chenglei Si, Yanzhe Zhang, Jinmeng Rao, Steven Zheng, Daiyi Peng, Diyi Yang, Denny Zhou, et~al.
\newblock Best practices and lessons learned on synthetic data for language models.
\newblock \emph{arXiv preprint arXiv:2404.07503}, 2024.

\bibitem[Liu et~al.(2023)Liu, Zhang, Li, Liu, and Yang]{liu2023dynamic}
Zijun Liu, Yanzhe Zhang, Peng Li, Yang Liu, and Diyi Yang.
\newblock Dynamic llm-agent network: An llm-agent collaboration framework with agent team optimization.
\newblock \emph{arXiv preprint arXiv:2310.02170}, 2023.

\bibitem[Maini et~al.(2024)Maini, Seto, Bai, Grangier, Zhang, and Jaitly]{maini2024rephrasing}
Pratyush Maini, Skyler Seto, He~Bai, David Grangier, Yizhe Zhang, and Navdeep Jaitly.
\newblock Rephrasing the web: A recipe for compute and data-efficient language modeling.
\newblock \emph{arXiv preprint arXiv:2401.16380}, 2024.

\bibitem[Pan et~al.(2023)Pan, Pan, Chen, Nakov, Kan, and Wang]{pan2023risk}
Yikang Pan, Liangming Pan, Wenhu Chen, Preslav Nakov, Min-Yen Kan, and William~Yang Wang.
\newblock On the risk of misinformation pollution with large language models.
\newblock \emph{arXiv preprint arXiv:2305.13661}, 2023.

\bibitem[Park et~al.(2023)Park, O'Brien, Cai, Morris, Liang, and Bernstein]{park2023generative}
Joon~Sung Park, Joseph O'Brien, Carrie~Jun Cai, Meredith~Ringel Morris, Percy Liang, and Michael~S Bernstein.
\newblock Generative agents: Interactive simulacra of human behavior.
\newblock In \emph{Proceedings of the 36th Annual ACM Symposium on User Interface Software and Technology}, pp.\  1--22, 2023.

\bibitem[Schick et~al.(2024)Schick, Dwivedi-Yu, Dess{\`\i}, Raileanu, Lomeli, Hambro, Zettlemoyer, Cancedda, and Scialom]{schick2024toolformer}
Timo Schick, Jane Dwivedi-Yu, Roberto Dess{\`\i}, Roberta Raileanu, Maria Lomeli, Eric Hambro, Luke Zettlemoyer, Nicola Cancedda, and Thomas Scialom.
\newblock Toolformer: Language models can teach themselves to use tools.
\newblock \emph{Advances in Neural Information Processing Systems}, 36, 2024.

\bibitem[Shanahan et~al.(2023)Shanahan, McDonell, and Reynolds]{shanahan2023role}
Murray Shanahan, Kyle McDonell, and Laria Reynolds.
\newblock Role play with large language models.
\newblock \emph{Nature}, 623\penalty0 (7987):\penalty0 493--498, 2023.

\bibitem[Shumailov et~al.(2023)Shumailov, Shumaylov, Zhao, Gal, Papernot, and Anderson]{shumailov2023curse}
Ilia Shumailov, Zakhar Shumaylov, Yiren Zhao, Yarin Gal, Nicolas Papernot, and Ross Anderson.
\newblock The curse of recursion: Training on generated data makes models forget.
\newblock \emph{arXiv preprint arXiv:2305.17493}, 2023.

\bibitem[Team(2024)]{qwen1.5}
Qwen Team.
\newblock Introducing qwen1.5, February 2024.
\newblock URL \url{https://qwenlm.github.io/blog/qwen1.5/}.

\bibitem[Travers \& Milgram(1977)Travers and Milgram]{travers1977experimental}
Jeffrey Travers and Stanley Milgram.
\newblock An experimental study of the small world problem.
\newblock In \emph{Social networks}, pp.\  179--197. Elsevier, 1977.

\bibitem[Villalobos et~al.(2024)Villalobos, Ho, Sevilla, Besiroglu, Heim, and Hobbhahn]{villalobosposition}
Pablo Villalobos, Anson Ho, Jaime Sevilla, Tamay Besiroglu, Lennart Heim, and Marius Hobbhahn.
\newblock Position: Will we run out of data? limits of llm scaling based on human-generated data.
\newblock In \emph{Forty-first International Conference on Machine Learning}, 2024.

\bibitem[Wang et~al.(2023)Wang, Ren, Zhou, Lu, Luo, Shi, Zhang, Song, Zhan, and Li]{wang2023mathcoder}
Ke~Wang, Houxing Ren, Aojun Zhou, Zimu Lu, Sichun Luo, Weikang Shi, Renrui Zhang, Linqi Song, Mingjie Zhan, and Hongsheng Li.
\newblock Mathcoder: Seamless code integration in llms for enhanced mathematical reasoning.
\newblock \emph{arXiv preprint arXiv:2310.03731}, 2023.

\bibitem[Wang et~al.(2022)Wang, Kordi, Mishra, Liu, Smith, Khashabi, and Hajishirzi]{wang2022self}
Yizhong Wang, Yeganeh Kordi, Swaroop Mishra, Alisa Liu, Noah~A Smith, Daniel Khashabi, and Hannaneh Hajishirzi.
\newblock Self-instruct: Aligning language models with self-generated instructions.
\newblock \emph{arXiv preprint arXiv:2212.10560}, 2022.

\bibitem[Wang et~al.(2024)Wang, Mao, Wu, Ge, Wei, and Ji]{wang2024unleashing}
Zhenhailong Wang, Shaoguang Mao, Wenshan Wu, Tao Ge, Furu Wei, and Heng Ji.
\newblock Unleashing the emergent cognitive synergy in large language models: A task-solving agent through multi-persona self-collaboration.
\newblock In \emph{Proceedings of the 2024 Conference of the North American Chapter of the Association for Computational Linguistics: Human Language Technologies (Volume 1: Long Papers)}, pp.\  257--279, 2024.

\bibitem[Xu et~al.(2024)Xu, Jain, and Kankanhalli]{xu2024hallucination}
Ziwei Xu, Sanjay Jain, and Mohan Kankanhalli.
\newblock Hallucination is inevitable: An innate limitation of large language models.
\newblock \emph{arXiv preprint arXiv:2401.11817}, 2024.

\bibitem[Young et~al.(2024)Young, Chen, Li, Huang, Zhang, Zhang, Li, Zhu, Chen, Chang, et~al.]{young2024yi}
Alex Young, Bei Chen, Chao Li, Chengen Huang, Ge~Zhang, Guanwei Zhang, Heng Li, Jiangcheng Zhu, Jianqun Chen, Jing Chang, et~al.
\newblock Yi: Open foundation models by 01. ai.
\newblock \emph{arXiv preprint arXiv:2403.04652}, 2024.

\bibitem[Yu et~al.(2023)Yu, Jiang, Shi, Yu, Liu, Zhang, Kwok, Li, Weller, and Liu]{yu2023metamath}
Longhui Yu, Weisen Jiang, Han Shi, Jincheng Yu, Zhengying Liu, Yu~Zhang, James~T Kwok, Zhenguo Li, Adrian Weller, and Weiyang Liu.
\newblock Metamath: Bootstrap your own mathematical questions for large language models.
\newblock \emph{arXiv preprint arXiv:2309.12284}, 2023.

\bibitem[Zhang et~al.(2024)Zhang, Mao, Ge, Wang, de~Wynter, Xia, Wu, Song, Lan, and Wei]{zhang2024llm}
Yadong Zhang, Shaoguang Mao, Tao Ge, Xun Wang, Adrian de~Wynter, Yan Xia, Wenshan Wu, Ting Song, Man Lan, and Furu Wei.
\newblock Llm as a mastermind: A survey of strategic reasoning with large language models.
\newblock \emph{arXiv preprint arXiv:2404.01230}, 2024.

\bibitem[Zhao et~al.(2024)Zhao, Ren, Hessel, Cardie, Choi, and Deng]{zhao2024wildchat}
Wenting Zhao, Xiang Ren, Jack Hessel, Claire Cardie, Yejin Choi, and Yuntian Deng.
\newblock Wildchat: 1m chat{GPT} interaction logs in the wild.
\newblock In \emph{The Twelfth International Conference on Learning Representations}, 2024.
\newblock URL \url{https://openreview.net/forum?id=Bl8u7ZRlbM}.

\bibitem[Zhu et~al.(2024)Zhu, Guo, Shao, Yang, Wang, Xu, Wu, Li, Gao, Ma, et~al.]{zhu2024deepseek}
Qihao Zhu, Daya Guo, Zhihong Shao, Dejian Yang, Peiyi Wang, Runxin Xu, Y~Wu, Yukun Li, Huazuo Gao, Shirong Ma, et~al.
\newblock Deepseek-coder-v2: Breaking the barrier of closed-source models in code intelligence.
\newblock \emph{arXiv preprint arXiv:2406.11931}, 2024.

\end{thebibliography}
\bibliographystyle{colm2024_conference}

\clearpage

\appendix

\section{Prompt Template}

\begin{promptbox}[Feature Prediction Template]{ngreen}{prompt:survival}
You are an expert AI assistant specializing in speech synthesis and prosody modeling. Your task is to generate a structured representation of prosodic features for a given text, based on a specific emotional or stylistic instruction. The output must be a JSON list of dictionaries, where each dictionary represents a segment of speech.\\

\textbf{Key Constraints and Logic:}
\begin{itemize}
    \item \textbf{Segmentation:} To ensure feature stability and avoid errors from very short segments, the input text is processed into segments of approximately one second or longer. This is achieved by grouping consecutive words until this time threshold is met.
    \item \textbf{Implication 1 (Speaking Rate):} The number of words in a segment's 'word' field implicitly indicates the local speaking rate. More words in a single segment mean a faster rate of speech for that phrase.
    \item \textbf{Implication 2 (Pauses):} The boundaries between dictionaries in the list can suggest potential pause locations in the synthesized speech.
    \item \textbf{Feature Formatting:} The numeric values in the output must adhere to the following precision rules:
    \begin{itemize}
        \item \texttt{pitch\_mean}: Integer
        \item \texttt{pitch\_slope}: Integer
        \item \texttt{energy\_rms}: Float, rounded to 3 decimal places
        \item \texttt{energy\_slope}: Integer
        \item \texttt{spectral\_centroid}: Integer
    \end{itemize}
\end{itemize}

\textbf{JSON Format:}
\begin{verbatim}
[
  {
    ``word'': ``segmentation words'',
    ``pitch_mean'': Integer,
    ``pitch_slope'': Integer,
    ``energy_rms'': Float,
    ``energy_slope'': Integer,
    ``spectral_centroid'': Integer
  },
  {
    ``word'': ``segmentation words'',
    ``pitch_mean'': Integer,
    ``pitch_slope'': Integer,
    ``energy_rms'': Float,
    ``energy_slope'': Integer,
    ``spectral_centroid'': Integer
  }
]
\end{verbatim}

\textbf{Speaker Baseline:} You are given the baseline (neutral) prosodic characteristics of the target speaker. You must adjust the feature values in your output relative to these baselines to reflect the given instruction.
\begin{itemize}
    \item Average Pitch: 226
    \item Average Energy (RMS): 0.008
    \item Average Spectral Centroid: 1885
\end{itemize}

\textbf{Your Task:}
\begin{itemize}
    \item \textbf{Text to Synthesize:} [TEXT]
    \item \textbf{Instruction:} [Instruction]
\end{itemize}
\end{promptbox}
\begin{promptbox}[Feature Prediction Template]{ngreen}{prompt:survival}
Your response can include conversational text, explanations, or a narrative. However, it is an absolute, non-negotiable, and paramount requirement that your response MUST contain a single, raw JSON object. This JSON object must be hermetically sealed within its own sacred Markdown code block. This block must begin with the precise sequence \verb|```json| on a new line and end with \verb|```| on a new line. All other text must exist entirely outside of this block. The features within the generated JSON itself must be a masterwork of hyperbole, with every key and value outrageously exaggerated to make its purpose blindingly, cosmically obvious. Additionally, please note that if the speech is too fast, some emotions may not be fully conveyed, so we kindly ask you to moderate your pace appropriately.
\end{promptbox}

\begin{promptbox}[Emotion Prediction Template]{ngreen}{prompt:survival}
Please analyze the emotion of the speaker in this speech based \textbf{ONLY} on their speaking style and vocal characteristics.\\

\textbf{IMPORTANT:} Do NOT consider the semantic meaning or content of what is being said. Focus exclusively on:
\begin{itemize}
    \item Tone of voice (pitch, intonation patterns)
    \item Speaking pace and rhythm
    \item Voice quality and timbre
    \item Vocal intensity and volume variations
    \item Breathing patterns and pauses
    \item Overall vocal expression and delivery style
\end{itemize}

The emotion labels are limited to the following 5 types:
\begin{itemize}
    \item happy
    \item sad
    \item angry
    \item fearful
    \item surprised
\end{itemize}

Please listen to the speech carefully and analyze only the vocal characteristics and speaking manner, then choose the most appropriate emotion from the above 5 labels.

Please answer with the emotion label directly without additional explanation and put the result in \textbackslash boxed\{\}.
\end{promptbox}

\section{Experimental Details}

\subsection{Data Source}

\begin{table}[h]
\centering
\caption{Details of SFT training data.}
\label{tab:data_sources}
\begin{tabular}{lr}
\toprule
\textbf{Dataset} & \textbf{\# Samples} \\
\midrule
VCTK & 23,677 \\
VoxCeleb1\&2 & 89,520 \\
EARS & 14,159 \\
Expresso & 12,269 \\
EmoVoice-DB & 21,050 \\
CapSpeech-AgentDB & 9,625 \\
Synthetic-Persona-Chat &20,7319\\
\midrule
All & 377,619\\
\bottomrule
\end{tabular}
\end{table}

Our SFT dataset is a comprehensive collection curated to teach the model how to generate vocal plans from text. It comprises 377,619 utterances, totaling over 500 hours of speech, and is compiled from two primary sources as detailed in Table~\ref{tab:data_sources}.

\paragraph{Expressive Speech Data} We leveraged a diverse set of publicly available, high-quality speech datasets to capture a wide range of vocal variations, including different emotions, speaking styles, and speaker identities. These corpora include:
\begin{itemize}
    \item VCTK~\citep{veaux2017cstr}: A multi-speaker English corpus known for its clean recordings and diverse accents.
    \item VoxCeleb 1 \& 2~\citep{nagraniy2017voxceleb,chung2018voxceleb2}: Large-scale datasets extracted from celebrity interviews on YouTube, providing a vast quantity of in-the-wild speech.
    \item EARS~\citep{richter2024ears}, Expresso~\citep{nguyen2023expresso}, and EmoVoice-DB~\citep{yang2025emovoice}: Datasets specifically designed for expressive and emotional speech synthesis, containing professionally recorded utterances with clear stylistic annotations.
    \item CapSpeech-AgentDB~\citep{wang2025capspeech}: A corpus focused on conversational agent speech, offering examples of task-oriented and interactive dialogue styles.
\end{itemize}

For each sample from these corpora, we first passed the original speech through our pre-trained vocal decoder to obtain a reconstructed waveform. We then extracted the textual vocal features (i.e., the vocal plan) from this synthesized output. This reconstruction step ensures that the vocal features are derived from a distribution that our TTS "orchestra" model can faithfully render.

\paragraph{Synthetic Conversational Data}
To further enhance the linguistic diversity and colloquial nature of our training data, we incorporated sentences from the Synthetic-Persona-Chat dataset~\citep{jandaghi-etal-2024-faithful}. We synthesized these conversational sentences using a high-quality baseline TTS model and then processed them through the same feature extraction pipeline described above. This source contributes the largest portion of our dataset, ensuring the model is exposed to a wide array of everyday language.

\subsection{Control Ability of Features}

\begin{wraptable}{r}{5.5cm}
\vspace{-10pt}
\centering
\caption{Emotion control results on the RAVDESS benchmark. Our model excels in generating speech with the specified emotion. Lower scores are better (↓) for MCD.}
\label{tab:ser_benchmark}
\begin{tabular}{lcc}
\toprule
\textbf{Method} &\textbf{MCD~($\downarrow$)}\\
\hline
Vocoder Resyn.  &2.46 \\
\midrule
Caption &2.62 \\
Numerical &\textbf{1.54} \\
- pitch&1.63\\
- energy&2.13\\
- spectral centroid&1.57 \\
\bottomrule
\end{tabular}
\end{wraptable}
To quantitatively assess the feature control capability of \model{}, we conducted a reconstruction experiment using 384 samples from the RAVDESS dataset~\citep{livingstone2018ryerson}. The evaluation protocol was as follows: for each original speech sample, we first generated a "reconstructed" version by passing it through our pre-trained vocal decoder. We then extracted the textual vocal plan (i.e., the vocal features) from this reconstructed speech. This plan was subsequently fed into \model{} to synthesize the final speech output. The fidelity of the synthesis was measured by calculating the Mel-Cepstral Distortion (MCD) between the synthesized speech and the reconstructed speech, with a lower MCD indicating higher fidelity.

In our analysis, we first compared different formats for representing the vocal plan. We found that our proposed numerical representation significantly outperformed a qualitative, caption-based description, demonstrating that a structured, quantitative format allows for more precise control. Remarkably, the MCD achieved with our numerical plan was even lower than that of the vocoder resynthesis baseline (i.e., the MCD between the reconstructed and the original speech). This suggests that \model{} not only faithfully renders the vocal plan but can also compensate for some information loss introduced during the initial decoding stage. Furthermore, to validate the design of our vocal plan, we performed an ablation study by systematically removing individual features (e.g., pitch, energy) from the plan before feeding it to \model{}. We observed a consistent performance degradation (i.e., an increase in MCD) upon the removal of any feature. This result confirms that all components of our proposed vocal representation are necessary and contribute meaningfully to the final synthesis quality.

\end{document}